\documentclass[a4paper,fleqn]{cas-dc}

\usepackage[numbers]{natbib}
\usepackage{enumitem}
\usepackage{graphicx}
\usepackage{caption}
\usepackage{subcaption}
\usepackage{mathtools}
\usepackage{adjustbox}
\usepackage{natbib}
\usepackage{amsmath}
\usepackage{multicol}

\DeclarePairedDelimiter\floor{\lfloor}{\rfloor}

\usepackage{nomencl}
\usepackage{framed}
\makenomenclature
\renewcommand*\nompreamble{\begin{multicols}{2}}
\renewcommand*\nompostamble{\end{multicols}}
\providetoggle{nomsort}
\settoggle{nomsort}{true} 

\makeatletter
\iftoggle{nomsort}{%
    \let\old@@@nomenclature=\@@@nomenclature        
        \newcounter{@nomcount} \setcounter{@nomcount}{0}%
        \renewcommand\the@nomcount{\two@digits{\value{@nomcount}}}
        \def\@@@nomenclature[#1]#2#3{
          \addtocounter{@nomcount}{1}%
        \def\@tempa{#2}\def\@tempb{#3}%
          \protected@write\@nomenclaturefile{}%
          {\string\nomenclatureentry{\the@nomcount\nom@verb\@tempa @[{\nom@verb\@tempa}]%
          \begingroup\nom@verb\@tempb\protect\nomeqref{\theequation}%
          |nompageref}{\thepage}}%
          \endgroup
          \@esphack}%
      }{}
\makeatother
\renewcommand{\printorcid}{}

\begin{document}\sloppy

\let\WriteBookmarks\relax
\def\floatpagepagefraction{1}
\def\textpagefraction{.001}
\shorttitle{Reinforcement learning for EV charging demand coordination}
\shortauthors{Lahariya M et~al.}
\setlength{\abovedisplayskip}{5pt} 
\setlength{\belowdisplayskip}{5pt} 
\setlength{\abovedisplayshortskip}{0pt}
\setlength{\belowdisplayshortskip}{0pt}
\captionsetup{justification=raggedright,singlelinecheck=false}
\newlist{romanlist}{enumerate*}{3}
\setlist[romanlist]{label=(\roman*)}
\newlist{arabiclist}{enumerate*}{3}
\setlist[arabiclist]{label=(\arabic*)}
\newenvironment{manumerate}
  {\itemize[noitemsep,label=$\bullet$]}             
  {\enditemize}
\def \questionname {Q}
\newcommand{\qref}[1]{\ref{#1}}  
\newlist{questionlist}{enumerate}{1}
\setlist[questionlist]{label=\textbf{(\questionname\arabic*)},ref=\questionname\arabic*,noitemsep,topsep=0pt,leftmargin=*}
  
\def \expname {Experiment}
\newcommand{\expref}[1]{\expname~\ref{#1}}
\newcounter{expcounter}
\newenvironment{experiment}[1]{\refstepcounter{expcounter}\textbf{\expname~\theexpcounter} #1}{}

\def\EV{EV}
\def\tslot{$\Delta t^\textit{slot}$}
\def\tdepart{$\Delta t^\textit{depart}$}
\def\tcharge{$\Delta t^\textit{charge}$}
\def\tflex{$\Delta t^\textit{flex}$}
\def\Tflex{$T_\textit{flex}$}
\def\Eflex{$E_\textit{flex}$}
\def\Smax{$\textit{S}_\textrm{max}$}
\def\Hmax{$H_\textrm{max}$}
\def\Nmax{$N_\textrm{max}$}
\def\Nflex{$N_d$}
\def\Ns{$N_s$}
\def\Cbau{$C_\text{BAU}$}
\def\Copt{$C_\text{opt}$}
\def\CRLQuad{$C_\textit{q}$}
\def\CRLAvg{$C^{E}_\textit{l,a}$}
\def\CRLMed{$C^{E}_\textit{l,m}$}
\def\mtot{$\textbf{m}^\textrm{total}_{s}$}
\def\action{$\textbf{u}_s$}
\def\actionflex{$\text{u}^d_s$}
\def\statex{$\textbf{x}_s$}
\def\matrixstate{$\textbf{x}_{s,m}$}
\def\vectorstate{$\textbf{x}_{s,v}$}
\def\actionspar{$\textbf{u}_{s,l}$}
\def\actionsful{$\textbf{u}_{s,g}$}
\def\actionsparflex{$\text{u}^d_{s,l}$}
\def\actionsfulflex{$\text{u}^d_{s,g}$}
\newcommand{\etc}{etc.\ }
\newcommand{\eg}{e.g., }
\newcommand{\Eg}{E.g., }
\newcommand{\ie}{i.e., }
\newcommand{\vs}{vs.\ }
\newcommand{\bigO}{\textit{O}}
\def \questionname {Q}
\def \figurename {Figure}
\def \Figurename {Figure}
\def \Sectionname {Section}
\def \Tablename {Table}
\def \Eqname {Eq.}
\def \appendixname {Appendix}
\def \algoname {Algorithm}
\def \experimentname {Exp}
\newcommand{\fref}[1]{\figurename~\ref{#1}}
\newcommand{\Fref}[1]{\Figurename~\ref{#1}}
\newcommand{\tref}[1]{\Tablename~\ref{#1}}
\newcommand{\sref}[1]{\Sectionname~\ref{#1}}
\newcommand{\aref}[1]{\appendixname~\ref{#1}}
\newcommand{\eref}[1]{\Eqname~(\ref{#1})}
\newcommand{\algoref}[1]{\algoname~\ref{#1}}

\title [mode = title]{
Computationally efficient joint coordination of multiple electric vehicle charging points using reinforcement learning
}
\author%
[1]{Manu Lahariya}
\ead[1]{manu.lahariya@ugent.be}

\author%
[2]{Nasrin Sadeghianpourhamami}
\ead[2]{nasrin.sadeghianpour@bluwave-ai.com}

\author%
[1]{Chris Develder}
\ead[1]{chris.develder@ugent.be}

\address[1]{IDLab, Ghent University -- imec, Technologiepark Zwijnaarde 126, 9052 Gent}
\address[2]{62 Steacie Drive, Suite 102, Ottawa, Ontario, Canada, K2K 2A9}

\begin{abstract}
A major challenge in today's power grid is to manage the increasing load from electric vehicle (EV) charging.
Demand response (DR) solutions aim to exploit flexibility therein, \ie the ability to shift EV charging in time and thus avoid excessive peaks or achieve better balancing.
Whereas the majority of existing research works either focus on control strategies for a single EV charger, or use a multi-step approach (\eg a first high level aggregate control decision step, followed by individual {\EV} control decisions), we rather propose a single-step solution that jointly coordinates multiple charging points at once.
In this paper, we further refine an initial proposal using reinforcement learning (RL), specifically addressing  computational challenges that would limit its deployment in practice.
More precisely, we design a new Markov decision process (MDP) formulation of the EV charging coordination process, exhibiting only linear space and time complexity (as opposed to the earlier quadratic space complexity).
We thus improve upon earlier state-of-the-art, demonstrating 30\% reduction of training time in our case study using real-world EV charging session data. Yet, we do not sacrifice the resulting performance in meeting the DR objectives: our new RL solutions still
improve the performance of charging demand coordination by 40-50\% compared to a business-as-usual policy (that charges EV fully upon arrival) and 20-30\% compared to a heuristic policy (that uniformly spreads individual {\EV} charging over time).
\end{abstract}


\maketitle

\begin{table*}[!t]   

\nomenclature{$s$}{State}
\nomenclature{$s'$}{The next state from $s$}
\nomenclature{$t$}{Timeslot}
\nomenclature{\tdepart}{Time left until departure}
\nomenclature{\tcharge}{Time needed for charging completion}
\nomenclature{\tflex}{Flexibility (time charging can be delayed)}
\nomenclature{\tslot}{Duration of a decision slot}
\nomenclature{$e$}{Episode}
\nomenclature{\Smax}{Maximum number of decision slots}
\nomenclature{\Hmax}{Maximum connection time}
\nomenclature{\Nmax}{Number of jointly coordinated charging stations}
\nomenclature{$N_s$}{Number of connected EVs in state $s$}
\nomenclature{$N_d$}{Number of EVs with flexibility \tflex $= d$}
\nomenclature{\statex}{Representation of state $s$}
\nomenclature{\matrixstate}{Aggregate demand in state $s$ (matrix state)}
\nomenclature{\vectorstate}{Aggregate flexibility in state $s$ (vector state)}
\nomenclature{\action}{Action taken in state $s$}
\nomenclature{\actionspar}{Locally scaled action taken in state $s$}
\nomenclature{${U}_{s,l}$}{Set of possible locally scaled actions from state $s$}
\nomenclature{\actionsful}{Globally scaled action taken in state $s$}
\nomenclature{${U}_{s,g}$}{Set of possible globally scaled actions from state $s$}
\nomenclature{\CRLQuad}{Quadratic cost function}
\nomenclature{\CRLAvg}{Linear cost function based on the average of preceding $E$ episodes}
\nomenclature{\CRLMed}{Linear cost function based on the median of preceding $E$ episodes}
\nomenclature{$L_{\pi}^{e}$}{Load in episode $e$ by policy $\pi$}
\nomenclature{$\tilde{L}_{\pi}^{e}$}{Normalized load in episode $e$ by policy $\pi$}
\nomenclature{$\mathcal{B}_{a}^{b}$}{EV session data from episode $a$ to episode $b$}
\nomenclature{$\mathcal{V}_{t}$}{Set of EVs in the system at time $t$}


\printnomenclature
\end{table*}
\section[Introduction]{Introduction}

Reliable operation of smart grids requires efficient load management, which can be achieved by demand response (DR) algorithms. A DR algorithm coordinates the demand-supply of energy to meet an objective, \eg load balancing, maximizing profit,~\etc Traditional DR is based on model predictive control (MPC)~\cite{Dkhili_2020_mpcsurvery,Donjol_2018_mpc}, where an optimization problem is solved using a predefined model. However, deployment of such model-based DR algorithms is limited due to uncertainties associated with the assumed models and lack of scalability and generalizability~\cite{nasrin_2016_modeling,Taha_2021_deepRLforDR}.

Model-free approaches circumvent the aforementioned challenges by formulating the problem using a Markov decision process (MDP) where the optimum policy is learned by interacting with the environment~\cite{nasrin2019journal,Wan_2019_EVRLscheduling,Hansen_2018_MDP,Otterlo_2012_MDP_RL}.
In recent years, data-driven reinforcement learning (RL) algorithms have been proposed based on an MDP for exploiting the user flexibility to coordinate the demand~\cite{nasrin2019journal,ONiell_2010_Q_learning}. 
In reinforcement learning, a coordinating agent learns by iteratively interacting with the environment and taking control actions.
The agent receives a reward/cost in each interaction and is thus trained to maximize/minimize the long term reward/cost.


The underlying MDP framework, which will be used in RL to find a (close to) optimal DR policy, is defined in terms of
\begin{romanlist}
    \item a $state$ representation, 
    \item an $action$ representation, and
    \item a reward/cost signal, \ie a $cost$ function
\end{romanlist}.
The specific DR objective (\eg peak shaving~\cite{Bijan_2022_peakshaving}, load balancing~\cite{nasrin2019journal, Manu_2019_RLcost}, valley filling~\cite{Chen_2012_valleyfitting}) is realized by appropriately designing the cost function, reflecting the utility of a certain action.
Yet, designing a meaningful but manageable MDP is challenging in practice. First of all, the complexity of the $state$-$action$ representation affects the space and computational complexity of the RL based control policy, thus potentially limiting its scalability (\eg computational requirements that depend on the number of EVs we aim to control jointly). Furthermore, an incorrect or uninformative cost function can degrade or annihilate the performance of the RL based control policy (known as the credit-assignment problem~\cite{Otterlo_2012_MDP_RL}). Thus, the MDP design can greatly impact the performance and optimization of an RL based control policy~\cite{Taha_2021_deepRLforDR}.

This paper defines and evaluates novel MDP formulations for {\EV} fleet charging coordination, which can be used to implement state-of-the-art RL-based DR. 
The proposed formulations represent a joint {\EV} coordination scenario, where we control  multiple {\EV} chargers jointly at once. This is more complicated than the common setting of controlling a single {\EV} charging point~\cite{Wan_2019_EVRLscheduling,Chis_2017_RLMDP,Shi_2011_MDP_EV_state}.
The proposed MDPs are defined using
\begin{romanlist}
 \item $state$-$action$ representations to facilitate scalability, and
 \item computationally linear cost functions to expedite learning
\end{romanlist}.
We will experimentally demonstrate the superiority of the newly engineered MDP in
terms of training time and space requirements
compared to the original MDP representation used in an
initial proof-of-concept paper~\cite{nasrin2019journal}.

In summary, the main contributions of this paper on jointly coordinating EV charging demand to reduce the resulting peak load include:
\begin{romanlist}
    \item the definition of novel $state$-$action$ representations based on aggregate demand and compact aggregate flexibility characterization (\sref{subsec:mdp}),
    \item the definition of novel linear cost functions that provide information on a posteriori established optimal {\EV} charging coordination over the recent past (\sref{sec:costfun}), and
    \item \label{it:contrib3} the quantitative analysis of the impact of the MDP formulation on the learning speed of the RL based control policy (\sref{sec:results}).
\end{romanlist}
For \ref{it:contrib3}, we provide experimental results using real-world EV charging data and the fitted $Q$-iteration (FQI) algorithm~\cite{Martin_2005_FQI} (with a neural network based function approximation, \sref{sec:batchRL}). 
Additionally, we compare our RL policies with both a business-as-usual policy (that charges EV fully upon arrival), and a heuristic policy (that uniformly spreads individual EV charging over time). In particular, we define experiments (\sref{sec:ExpDetails}) to answer the following questions:
\begin{questionlist}
    \item \label{q:stateactionfeatures} What features (\eg connection time, required energy, etc.) should be used to define $state$-$action$ representations? 
    \item \label{q:costfunctiondef} 
	What is the impact of the cost function definition on the learning speed and performance of the learned policy?
	\item \label{q:impactofmdp} What is the impact of our MDP formulation on
    	\begin{romanlist}
    	    \item the performance of the RL policy, and
    	    \item the space and computation time complexity of learning such optimum RL policy?
    	\end{romanlist}
	\item \label{q:parametersettings} What is the impact of varying parameter settings during training on the performance of the RL policy?\footnote{The parameters of interest are
	\begin{romanlist}
	    \item the number of preceding days used in cost function,
	    \item the time span of the training data, and
	    \item the number of iterations in FQI algorithm.
	\end{romanlist}
	For details see~\sref{sec:Experiments} and \sref{sec:results_crd}.}
\end{questionlist}
Conclusions and open issues for future work are presented in \sref{sec:conclusions}.


\section[Related work]{Related Work}

In recent years, RL has attracted attention to facilitate data-driven DR to coordinate {\EV} charging. Researchers have exploited RL for charging coordination of individual {\EV}s to achieve objectives such as cost reduction, providing customized services, \etc For example, \citet{Chis_2017_RLMDP} achieve 10\%-50\% reduction in long term cost compared to business-as-usual using a RL control policy with a MDP framework where
\begin{romanlist}
    \item the states are based on charging prices, price fluctuations, and timing variables, and 
    \item the actions are based on daily energy consumption
\end{romanlist}. 
Controlling the charging/discharging of an individual EV under price uncertainty for providing vehicle-to-grid (V2G) services is studied by~\citet{Shi_2011_MDP_EV_state}. Their MDP has 
\begin{romanlist}
    \item a state based on the hourly electricity price, state-of-charge, and time left till departure, and
    \item an action based on the decision of charging, delaying the charging, and discharging
\end{romanlist}.
Cost-saving by scheduling charging/discharging of a single {\EV} for a user is studied by~\citet{Wan_2019_EVRLscheduling}, where the MDP is based on 
\begin{romanlist}
    \item a state defined using electricity price and current state of charge of {\EV} and,
    \item a binary action of either charging or discharging 
\end{romanlist}.
As opposed to the previously stated cases~\cite{Chis_2017_RLMDP,Shi_2011_MDP_EV_state,Wan_2019_EVRLscheduling}, where the authors study controlled charging of a \emph{single} {\EV}, we study the joint coordination of a \emph{group} of {\EV}s. Additionally, the objective of our control policy is load flattening in contrast to the objective of cost-minimization. 

For the collective charging of a \emph{group} of {\EV}s, previous studies have proposed stepwise approaches. \citet{Vandael_2013_peakshaving} propose a three-step DR algorithm for a group of {\EV}s, based on which~\citet{Claessens_2013_peakshaving} learn a collective charging plan for {\EV}s using batch RL. Both their three-step DR approaches comprise an aggregation step, an optimization step, and a real-time control step. A heuristic algorithm is used in the control step to dispatch the energy corresponding to the charging actions determined in the optimization step. Batch RL is employed to generate these actions in the optimization step and trained using the aggregated constraints for each {\EV}. 
A two-step time-of-day pricing demand response is proposed in~\cite{Jose_2021_EVpricingstrategy}, where an aggregator agent sells energy to clusters of EVs. Interactions in these aggregator agents is formulated as a two-step optimization problem, where, 
\begin{romanlist}
    \item at the top level the aggregator maximizes its benefits, and,
    \item at the lower level the rational of EV driver is optimized  
\end{romanlist}.
Contrary to these stepwise DR approaches based on separate aggregation and optimization steps, our DR jointly controls all {\EV}s directly based on an efficient aggregated state representation.

The current paper more specifically addresses real-world implementation and scalability challenges of RL-based control, by defining and exploring state-of-the-art MDP formulations. In our previous work~\cite{nasrin2019journal}, we provided a proof-of-concept for RL-based DR for joint {\EV} coordination (with MDP formulation that has quadratic cost function and $state$-$action$ representation). In the current work we further focus on the MDP formulation (which we refine) and the learning speed of said RL-based DR.
More specifically, in this paper, we,
\begin{romanlist}
    \item define new compact $state$-$action$ representations that scale linearly with system capacity and coordination horizon in contrast to exponential scaling in~\cite{nasrin2019journal},
    \item propose computationally linear cost functions compared to the quadratic cost function in~\cite{nasrin2019journal}, and 
    \item we study the impact of our cost functions on RL based control policy optimization (by evaluating the computation time per iteration in FQI algorithm), in contrast, no such analysis is done in~\cite{nasrin2019journal}.
\end{romanlist}
In the next section, we define the different components of the MDP.

\section{Markov Decision Process (MDP)}
\label{subsec:mdp}

Our main focus is flattening the load stemming from charging a group of {\EV}s, \ie we aim to minimize the peak-to-average ratio of the aggregate charging load of the group of EVs. This objective is achieved by minimizing the long term cost or T-step return in a Markov decision process (MDP). We use the MDP framework for model-free coordination for aggregated electric vehicle charging, by formulating the charging coordination as a discrete sequential decision making process. The learning objective is to find an optimum coordinated charging policy such that the long term expected cost is minimized. 

We use RL to learn an optimum policy that achieves the said load flattening objective by jointly coordinating the charging demand of {\EV}s connected to {\Nmax} charging stations. We assume that state transition probabilities are a priori unknown, thus, the policy is learned from interacting with the environment, by taking actions and observing the rewards/costs and next state. In the following subsections, we define two $state$ representations, two $action$ representations, and different cost functions that can be used to define the MDP formulation.

\subsection{State Representations}
\label{sec:state}

An individual EV charging session is characterized by the
\begin{romanlist}
    \item EV arrival time,
    \item EV departure time ({\tdepart}),
    \item required energy, and 
    \item charging power
\end{romanlist}. We implicitly assume the same charging power for all EVs and compute the time needed to complete the charging ({\tcharge}) by dividing the required energy with the charging power. A \textit{state} representation is defined using the information from these features. 
At timesetp $t$, the number of {\EV}s in the system is {\Ns}, and the available information can be summarized as:
\begin{equation}
    \mathcal{V}_t = \{(\Delta t ^{depart} _ {1}, \Delta t ^{charge} _ {1}), \dots, (\Delta t ^{depart} _ {N_{s}}, \Delta t ^{charge} _ {N_{s}})\} 
    \label{eqn:matrixsetinfo}
\end{equation}
The underlying idea of charging demand coordination is to exploit the available flexibility in the system. Thus, this flexibility can be used to define the state representation. The \textit{flexibility}, \ie how much charging can be delayed, offered by each EV is represented by {\tflex} = {\tdepart} $-$ {\tcharge}. We thus redefine the avaliable information as:
\begin{equation}
        \mathcal{V}'_t = \{(\Delta t ^{flex} _ {1}), \dots, (\Delta t ^{flex} _ {N_{s}})\}
    \label{eqn:vectorsetinfo}
\end{equation}
The state representation is of the form $s=(t,$ \statex), where $t$ is the timeslot ($\ie t \in \{1, \dots, $ \Smax$\}$)
and {\statex} provides the aggregate information of all EVs in the system. A binning algorithm inspired by~\cite{Claessens_2016_binning} is used to aggregate the EVs connected to the charging stations while retaining the required information to facilitate learning.

We previously~\cite{nasrin2019journal} defined a matrix state representation {\matrixstate} using set $\mathcal{V}_t$. We now propose a new compact and improved vector state representation {\vectorstate} using set $\mathcal{V}'_t$.

\begin{enumerate}[label={(\arabic*)}]
    \item \textit{\textbf{Matrix State Representation }}({\matrixstate}): The aggregate demand represented in a 2D grid, with one axis representing {\tdepart}, and other {\tcharge}. The element of {\matrixstate} at position $(i, j)$ counts the number of {\EV}s in the corresponding ({\tdepart}, {\tcharge}) bin, \ie for which $i$ = {\tdepart} and $j$ = {\tcharge}. 
    
    \item \textit{\textbf{Vector State Representation}} ({\vectorstate}): The aggregate flexibility represented using a vector. The element of {\vectorstate} at position $i$ counts the number of {\EV}s in the corresponding ({\tflex}) bin, \ie for which $i$ = {\tflex}. 
\end{enumerate}

\begin{figure*}[pos = !tb, ,align=\raggedright]
\centering
\begin{subfigure}{.45\textwidth}
  \centering
  \includegraphics[width=0.99\linewidth]{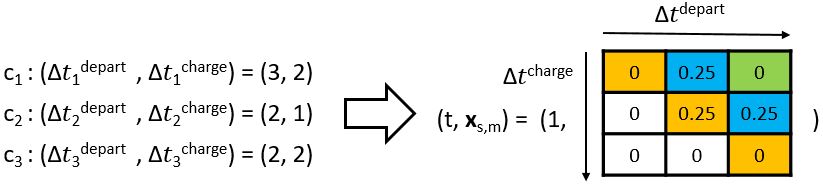}
  \caption{Fully observable matrix state representation ({\matrixstate}) that provides aggregate demand.}
  \label{fig:matrix_rep}
\end{subfigure}%
\hfill
\begin{subfigure}{.5\textwidth}
  \centering
  \includegraphics[width=0.77\linewidth]{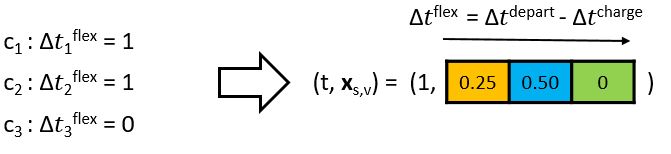}
  \caption{Partially observable vector state representation ({\vectorstate}) that provides aggregate flexibility.}
  \label{fig:vector_rep}
\end{subfigure}
\caption{Aggregate representations of state $s$ with system capacity {\Nmax} = 4, and decision slots {\Smax} = 3. Each state is characterized by timeslot $t$ and an aggreate state representation, for example, $s$ = ($t$,\matrixstate). (Yellow: {\tflex}=0; Blue: {\tflex}=1; Green: {\tflex}=2)}
\label{fig:state_rep}
\end{figure*}

The maximum number of decision slots is {\Smax} = {\Hmax}/{\tslot} ({\Hmax} is the maximum connection time). For an exemplary scenario with {\Nmax} = 4 and {\Smax} = 3, \fref{fig:matrix_rep} provides the matrix state representation ({\matrixstate}) and \fref{fig:vector_rep} provides the vector state representation ({\vectorstate}). At time $t$~= 1, assume that we have $N_s$ = 3 cars: $c_1$~= ({\tdepart}, {\tcharge})~= (3,2), $c_2$~= ({\tdepart}, {\tcharge})~= (2,1) and $c_2$~= ({\tdepart}, {\tcharge})~= (2,2) with no other arrivals during the control horizon. 
The resulting states are divided by {\Nmax}~= 4 to maintain generalizability in our state representations. We have shown previously that the policy learned from the {\matrixstate} representation can be generalized to a different number of charging stations if the state representation is normalized by system capacity {\Nmax}. To maintain this generalizability in our new state representations, we perform the same normalization.

In the matrix state representation, {\EV}s binned into cells on the main diagonal of {\matrixstate} (i.e., $i$ = $j$) have zero flexibility while
the ones binned into cells on the upper diagonals of {\matrixstate} are
flexible charging requests. In a vector state representation EVs binned into the first cell of {\vectorstate} (i.e., $i$ = 0) have zero flexibility while
the ones binned into cells after the first cell of {\vectorstate} are
flexible charging requests. Negative {\tflex}, corresponding to the lower diagonals of {\matrixstate}, would indicate {\EV}s for which the requested charging demand cannot be fulfilled (\tcharge > \tdepart).

The matrix state representation summarizes all the information available from the environment and results in a fully observable setting. The information about the flexibility ({\tflex} = {\tdepart}$-${\tcharge}) can also be identified from the matrix state representation. The vector state representation only summarizes the information about flexibility, and results in a partially observable setting. Yet it still is highly relevant for making charging decisions. We use the matrix and vector state representations to answer~\qref{q:stateactionfeatures} (What features should be used to define $state$-$action$ representations?), where these representations are based on different features and provide different levels of detail, \ie observability, to learn the control policy. Note that the vector state representation results in a smaller number of states in the state space of the problem compared to the matrix state representation.

\subsection{Action Representations} 
\label{sec:action}

The action our agent needs to decide on is
which EVs to charge and which ones' charging to delay in state $s$ = (t,~{\statex}). We will make the decisions based on flexibility ({\tflex}): {\EV}s that offer similar flexibility will be charged/delayed together. Thus, actions are represented by a vector {\action} where the element at position $d$, given by {\actionflex}, provides the number of {\EV}s to charge in flexibility bin {\tflex} = $d$ ($d \in \{0, \dots,$ \Smax $-$ 1\}). The total number of {\EV}s in flexibility bin {\tflex} = $d$ is {\Nflex}. Thus, the element {\actionflex} of the action vector will be a number in \{0, \dots, {\Nflex}\}. For example, {\Nflex} = 3 means that 3 cars offer the same flexibility, and {\actionflex} is the number of cars that will be charged, and thus lies in \{0, 1, 2, 3\}.

We scale the elements of action {\action} to be numbers in [0,1], representing the fraction of cars that we will charge, \ie we divide {\action} by {\Nflex} or {\Nmax}.
We previously~\cite{nasrin2019journal} divided action {\action} by {\Nflex} to estimate a \emph{locally} scaled action {\actionspar}, which is difficult to interpret as {\Nflex} depends on $d$ and changes after each interaction with the system. To improve interpretability, we now divide action {\action} by {\Nmax} to estimate a \emph{globally} scaled action {\actionsful}, and keeping the scaling factor fixed.
We thus have the following two action representations:

\begin{enumerate}[label={(\arabic*)}]
\item \textit{\textbf{Locally Scaled Action Representation }}({\actionspar}): Each element of the action vector is divided by the total number of cars in its flexibility bin, \ie {\actionsparflex} = {\actionflex}/{\Nflex}.  

\item \textit{\textbf{Globally Scaled Action Representation }}({\actionsful}):  Each element of the action vector is divided by the system capacity ({\Nmax}), \ie {\actionsfulflex} = {\actionflex}/{\Nmax}. 
\end{enumerate}

\begin{figure*}[pos = !tb, ,align=\raggedright]
\centering
   \includegraphics[width=0.8\linewidth]{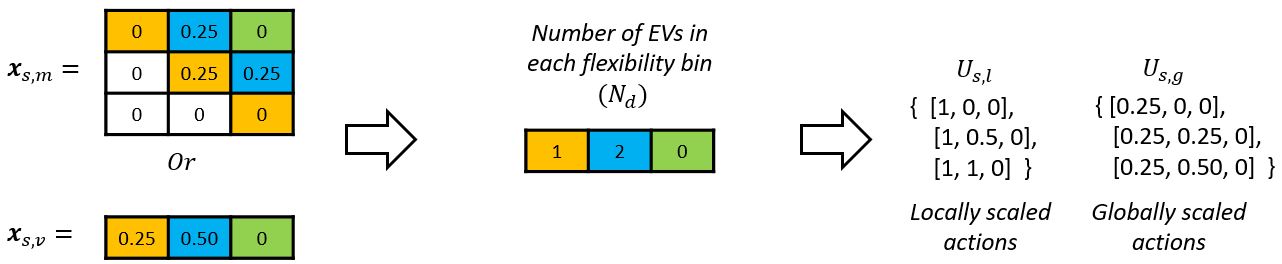}
    \caption{Action spaces for different state representations ($U_{s,l}$: Action space with locally scaled actions, $U_{s,g}$: Action space with globally scaled actions). Derived from the matrix (\matrixstate) or vector (\vectorstate) state representations for a system with {\Nmax} = 4 {\EV} charging points. (Yellow: {\tflex}=0; Blue: {\tflex}=1; Green: {\tflex}=2)}
    \label{fig:action_spaces} 
\end{figure*}

For the exemplary scenario stated before with {\Nmax}~= 4 and $N_s$~= 3,
\fref{fig:action_spaces} illustrates how {\actionspar} and {\actionsful} are constructed from both state representations ({\matrixstate} and {\vectorstate}) and the counts {\Nflex}. We will ensure that {\EV} charging demands are never violated, \ie the {\EV}s for which {\tflex} = 0 will always be charged. Thus, there are 3 possible actions to take in this state. For the other two {\EV}s the possible actions are: charge none, charge one or charge both. State representations are used to estimate an intermediate vector in which the value at position $d$ represents {\Nflex}, \ie the total number of {\EV}s with flexibility $d$. This intermediate vector can be used to define action space $U_{s,l}$ for locally scaled actions and action space $U_{s,g}$ for globally scaled actions.

\subsection{Cost Function} 
\label{sec:costfun}

Our objective is to flatten the aggregate {\EV} charging load, while fulfilling the charging demand for each EV before its departure. This will be achieved through defining a cost function denoted as $C(s,$\action$,s')$ quantifying the utility of a transition from state $s$ to $s$' by taking action {\action}. Each state $s$ is given by ($t$, \statex), where $t$ is the timeslot and {\statex} is the state (matrix or vector state representation). 
The cost function will clearly be related to charging load, \ie power consumption. Power consumed from all {\EV}s by taking action {\action} in state {\statex} will be represented by $P (\textbf{x}_s,\textbf{u}_{s})$:
\begin{equation}
    P (\textbf{x}_s,\textbf{u}_{s}) = \sum\limits^{S_{max}-1}_{d=0} N_d \textbf{u}_{s}^d
    \label{eqn:policypowerconsumption}
\end{equation}
\subsubsection{{Quadratic Cost Function} ({\CRLQuad})} 
In our previous work~\cite{Manu_2019_RLcost}, we defined a cost function based on the squared power consumption. This quadratic cost function is defined in \eref{eq:RL_quadcost} where the cost for taking action {\action} in state $s$ =$(t,$\statex$)$ amounts to the squared charging power consumption. 
\begin{align}
    \textit{C}_q(s, \textbf{u}_{s}, s') &= \textit{C}_q(\underbrace{(t,\textbf{x}_s)}_{s},\textbf{u}_{s},\underbrace{(t+1,\textbf{x}_{s'})}_{s'}) \nonumber \\
    &\triangleq\left(P(\textbf{x}_s,\textbf{u}_{s})\right)^2
\label{eq:RL_quadcost} 
\end{align}
Minimizing the long term cost based on this function translates to load-flattening across all {\EV} charging stations. 
\subsubsection{{Linear Cost Functions} ({\CRLAvg} and {\CRLMed})}
 We define cost functions that utilize the information from optimum policies of the preceding days (where we have information for all {\EV} sessions). This is based on the assumption that days with similar {\EV} session characteristics (arrivals, departures and required energy) will have similar optimal solutions.
 The optimal solution, \ie power consumption for each timeslot, for a prior day can be calculated using an all-knowing optimum policy (\eg by formulating the problem as a quadratic optimization problem). The power consumed from all {\EV}s under optimal policy coordination is represented by $P^{opt} (t,e)$, where $t$ is timeslot and $e$ represents the corresponding episodic `day'.\footnote{
We will assume that {\EV} charging stations empty at the end of a day. Thus, an episodic `day' constitutes the period during which {\EV}s are present, and ends with an empty car park at the end of the day.
For details see~\sref{sec:training}}
For the set of preceding $E$ episodic `days', the consumed powers from episode $e-1$ to $e-E$ can be summarized in the set defined in~\eref{eqn:setpastEopt}, where $e$ is the current episodic `day'.
\begin{equation}
    \mathcal{P}^{opt}_{t,e,E} = \{P^{opt} (t,e-1),\dots,P^{opt} (t,e-E) \}
    \label{eqn:setpastEopt}
\end{equation}
By utilizing the information about the optimal policy available in this set, we define two cost functions. These cost functions are based on the absolute difference between power consumed, \ie $P^{\pi}(\textbf{x}_s,\textbf{u}_{s})$, and either the average (\eref{eq:RL_linavgcost}) or the median (\eref{eq:RL_linmedcost}) of set $\mathcal{P}^{opt}_{t,e,E}$. 
\begin{equation}
    \textit{C}_{l,a}^E(s, \textbf{u}_{s}, s') \triangleq \left| P^{\pi}(\textbf{x}_s,\textbf{u}_{s}) - \text{avg}(\mathcal{P}^{opt}_{t,e,E}) \right|
\label{eq:RL_linavgcost} 
\end{equation}
\begin{equation}
    \textit{C}_{l,m}^E(s, \textbf{u}_{s}, s') \triangleq \left| P^{\pi}(\textbf{x}_s,\textbf{u}_{s}) - \text{median}(\mathcal{P}^{opt}_{t,e,E}) \right|
\label{eq:RL_linmedcost} 
\end{equation}
The second term of these cost functions is the average/median of power consumption by optimum policy coordination for the preceding $E$ episodic `days'. Minimizing the long term cost calculated from these cost functions translates to reducing the deviation of the current charging policy ($\pi$) from the optimum policies of preceding episodes. Hence, the current charging policy ($\pi$) learns to mimic the behavior, and subsequently the objective, of these optimum policies. These cost functions will be effective for any coordination objective (\eg cost-saving, peak-shaving, \etc), as the current charging policy approximates the optimum policies of preceding episodes, which can be trained for any objective. We consider the case of load-flattening to test these linear cost functions against the quadratic cost function to answer~\qref{q:costfunctiondef} (Impact of the cost function definition on the learning speed and performance of the learned policy?). 

\subsection{State-Action Value Function} 

Solving the {MDP} means finding an optimum control policy 
$\pi$ : $S \to U$ that minimizes the expected $T$-step cost, which for a policy $\pi$ at timestep $t$ is given in \eref{eq:RL:Ereturn}.   The control time horizon in our setting is given by $T=$ \Smax. The policy can be identified by evaluating a state-action value function, \ie the $Q$-function, and selecting the action that minimizes it at each timestep. This $Q$-function is provided in \eref{eq:RL:Qfunc}.
\begin{equation}\label{eq:RL:Ereturn}
J^{\pi}_T(s) =\mathbb{E} \left[\sum_{i=t}^{t+T}{C(\underbrace{(t,\textbf{x}_s)}_{s},\textbf{u}_{s},\underbrace{(t+1,\textbf{x}_{s'})}_{s'})}\right]
\end{equation}
\begin{equation}\label{eq:RL:Qfunc}
Q^{\pi}(s,\textbf{u}_s) =\mathbb{E} \left[C(s,\textbf{u}_s,s')+J^{\pi}_T(s')\right].
\end{equation}
The $Q$-function corresponding to the optimum policy represented by $Q^{*}(s,\textbf{u}_s)$  = $\min_{\pi}Q^{\pi}(s,\textbf{u}_s)$ can be calculated if the transition probabilities between states are known. However, these are unknown in our setting, hence we use a learning algorithm to approximate the optimum $Q$-function as $\widehat{Q}^{*}(s,\textbf{u}_s)$. In the next section, we provide details of the batch reinforcement learning algorithm used to learn $\widehat{Q}^{*}(s,\textbf{u}_s)$.

\section{Batch Reinforcement Learning}
\label{sec:batchRL}

We employ batch mode reinforcement learning to learn the $Q$-function, \ie $\widehat{Q}^{*}(s,\textbf{u}_s)$. In batch reinforcement learning algorithms, optimization is performed on data collected in past experiences rather than online interactions from the environment. We use the historical EV data (arrivals, departures, and required energy) and a random policy to collect past experiences. Each experience is defined in terms of
\begin{romanlist}
    \item an initial state $s$,
    \item the action taken {\action},
    \item the resulting state $s'$ after taking the action, and
    \item the associated costs $C(s,$\action$,s')$
\end{romanlist}.
An experience set denoted by $\mathcal{F}$ contains tuples $(s,$\action$,s',C(s,$\action$,s'))$ and is generated based on the state representation, action representation, and cost function. 

\paragraph{{Fitted Q-iteration (FQI):}}
{We use the Fitted Q-iteration~\cite{Martin_2005_FQI} algorithm to learn $\widehat{Q}^{*}(s,\textbf{u}_s)$ from $\mathcal{F}$.
A fully connected Artificial Neural network (ANN) is used as function approximation for $\widehat{Q}^*(s,\textbf{u}_s)$.}

\paragraph{{State-Action space}:}
{The past experience set $\mathcal{F}$ is generated by taking actions (randomly or deterministically) given a state $s$. The number of all possible actions given a state $s$ is given by~\eref{eq:RL:stateacspace}, where the second term reflects the single ``choice'' we have for the cars without flexibility ({\tflex} = 0).
\begin{equation}\label{eq:RL:stateacspace}
\left|\textbf{U}_{s}\right| = \prod_{d=1}^{S_{\textit{max}}-1}\left(N_d + 1 \right).
\end{equation}
}

\section[Experiment Setup]{Experiment Setup}
\label{sec:ExpDetails}

In this section, we provide the details of the real-world data used in our experiments. We outline the experiments that were performed to answer our research questions. We also provide performance metrics used to evaluate the trained control policies. 

\begin{table*}[pos = !tb, ,align=\raggedright]
    \begin{tabular*}{0.99\textwidth}{@{\extracolsep{\fill}}c|c|c|c|c|c}
      {Experience} & {State} & {Action} & {Cost Function} & {RL Trained} & {State Space} \\
      {Set} & {Representation} & {Representation} & {} & {Policy} & {Complexity} \\
      \hline
      $\mathcal{F}_1$ & Matrix (t, {\matrixstate}) & Locally Scaled ({\actionspar}) & Quadratic ({\CRLQuad}) & RL$_{ml}$ & $\bigO(S_{max}^2)$ \\
      $\mathcal{F}_2$ & Vector (t, {\vectorstate}) & Locally Scaled ({\actionspar}) & Quadratic ({\CRLQuad})& RL$_{vl}$ & $\bigO(S_{max})$\\
      $\mathcal{F}_3$ & Matrix (t, {\matrixstate}) & Globally Scaled ({\actionsful}) & Quadratic ({\CRLQuad})& RL$_{mg}$ & $\bigO(S_{max}^2)$\\
      $\mathcal{F}_4$ & Vector (t, {\vectorstate}) & Globally Scaled ({\actionsful}) & Quadratic ({\CRLQuad})& RL$_{vg}$ & $\bigO(S_{max})$
    \end{tabular*}
    \caption{Experience sets ($\mathcal{F}$) generated for~\expref{exp:obs} (observability) to evaluate the effect of information provided by $state$-$action$ representation on the performance of learned RL control policy.}
    \label{table:SArepExpsets}
\end{table*}
\begin{table*}[pos = !tb, ,align=\raggedright]
    \begin{tabular*}{0.99\textwidth}{@{\extracolsep{\fill}}c|c|c|c|c|c}
      {Experience} & {State} & {Action} & {Cost Function} & {RL Trained} & {State Space} \\
      {Set} & {Representation} & {Representation} & {} & {Policy} & {Complexity} \\
      \hline
      $\mathcal{F}_5$ & Vector (t, {\vectorstate}) & Globally Scaled ({\actionsful}) & Quadratic ({\CRLQuad})& RL$_{q}$ & $\bigO(S_{max})$\\
      $\mathcal{F}_6$ & Vector (t, {\vectorstate}) & Globally Scaled ({\actionsful}) & Linear (Average) ({\CRLAvg})& RL$_{a}$ & $\bigO(S_{max})$\\
      $\mathcal{F}_7$ & Vector (t, {\vectorstate}) & Globally Scaled ({\actionsful}) & Linear (Median) ({\CRLMed})& RL$_{m}$ & $\bigO(S_{max})$
    \end{tabular*}
    \caption{Experience sets ($\mathcal{F}$) generated for~\expref{exp:crdassignment} (credit-assignment) to evaluate the effect of different cost functions on the performance of learned RL control policy.}
    \label{table:CAExpSets}
\end{table*}

\subsection{Data and Model Specifications}
\label{sec:training}

We use a real-world dataset to train the RL based control policy. Our dataset is derived from real-world data collected by ElaadNL since 2011, from 2500+ public charging stations \cite{nasrin2018data}, from which we selected the data for 2015. We represent this data in an episodic format, such that each episodic `day' starts at 7 am and ends 24 hours later (the day after at 7 am). Further, we assume an empty car park at end of each episode (all {\EV}s leave the charging stations). Thus, irrespective of the starting state of the episode, we always reach the same terminal state characterized by an empty car park (all elements of the {\matrixstate} or {\vectorstate} are zero). A terminal state stabilizes the learning process in FQI that adopts a neural network based function approximation~\cite{Riedmiller_2012_tipsRL}.
The time granularity is set to {\tslot} = 2~h, which means {\Smax} = 12 timeslots in each episodic `day'.
We jointly coordinate {\Nmax} = 10 charging stations,\footnote{We select the 10 busiest {\EV} charging stations (based on the total number of transactions).} 
\ie at most 10 EVs can be connected simultaneously.

For training the RL agent we start by creating the experience sets $\mathcal{F}$. This set contains the past experience for multiple episodes. For each episode, we start from the first state of a day characterized by ($t_1, x_1$) and randomly choose an action from the set of possible actions in each state and observe the next state and the associated state transition cost until the terminal state is reached (i.e., ($t_T, x_T$)). This single sequence of states and actions is referred to as a trajectory. Each transition in this trajectory is saved in the experience set in the form of tuples $(s,$\action$,s',C(s,$\action$,s'))$. We randomly generate 5000 trajectories for each episode.

An artificial neural network (ANN) architecture is used to estimate the $Q$-function from the experience set $\mathcal{F}$ using FQI. This network consists of an input layer and 2 hidden layers with ReLU activation functions.  There are
128 and 64 neurons in the first and second hidden layers
respectively. The output layer has a single neuron with a linear activation function. The input of the network is a vector that is created by combining the state and action. In case of matrix state representation we concatenate the matrix rows to obtain an input vector of length
145 (({\Smax})$^2$ + {\Smax} + 1), and in the case of vector state representation, the length of the input vector is 25 ({\Smax} + {\Smax} + 1). We use Huber
loss~\cite{Peter_1964_robust} instead of mean-squared-error for improving the stability in learning in our algorithm~\cite{Mnih_2015_huberloss}. All experiments are run on a system with an Intel Xeon E5645 3.1\,GHz processor and 16\,GB RAM. We outline two experiments below that are used to answer our research questions.

\begin{figure*}[pos = !tb, ,align=\raggedright]
\centering
\begin{subfigure}{.5\textwidth}
  \centering
  \includegraphics[width=0.7\linewidth]{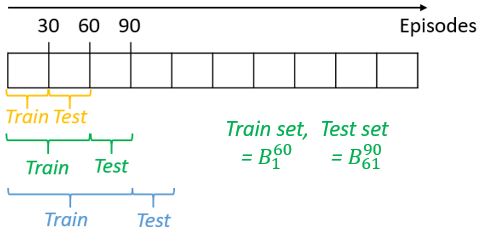}
  \caption{Increasing window validation used in~\expref{exp:obs}}
  \label{fig:increasingwindow}
\end{subfigure}%
\begin{subfigure}{.5\textwidth}
  \centering
  \includegraphics[width=0.7\linewidth]{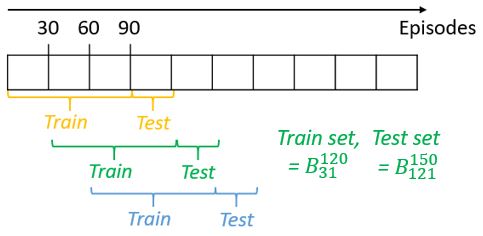}
  \caption{Rolling window validation used in~\expref{exp:crdassignment}}
  \label{fig:rollingwindow}
\end{subfigure}
\caption{Train and test data selected in different validation methods used in experiments}
\end{figure*}

\subsection{Experiments}
\label{sec:Experiments}

We define two experiments utilizing the ElaadNL transactions data of 2015 to answer our research questions. A dataset of transactions from episode $a$ to $b$ is represented by $\mathcal{B}_{a}^{b}$. A validation set thus contains a training dataset $\mathcal{B}_{a}^{b}$ and a testing dataset $\mathcal{B}_{c}^{d}$. We evaluate the performance of RL based control policies across all validation sets.

\paragraph{\begin{experiment}\label{exp:obs} Observability:\end{experiment}}
We design an experiment to evaluate the impact of different $state$-$action$ representations and answer~\qref{q:impactofmdp} (Impact of MDP formulation on the performance of the RL policy?) and a part of~\qref{q:parametersettings} (What is the impact of varying parameter settings during training the RL agent?).
Based on different $state$-$action$ representations defined in \sref{sec:state} and \sref{sec:action}, we generate 4 different experience sets. The details of these experience sets are summarized in~\tref{table:SArepExpsets}. Each experience set is used to learn a control policy. For example, RL$_{ml}$ is a trained control policy from experience set $\mathcal{F}_1$ which has matrix state representation and locally scaled action representation. 

The $state$-$action$ representations affect the space complexity of an experience set ($\mathcal{F}$), and the learning speed of the control policy. To compare different $state$-$action$ representations (\tref{table:SArepExpsets}), we perform an increasing window validation where the size of training datasets are different (\fref{fig:increasingwindow}). The training datasets are generated from $\{30, 60,\dots,270\}$ episodes, and we test on the immediate next 30 episodes. For example, the first validation set has training data $\mathcal{B}_{1}^{30}$ (data from \{1, \dots, 30\} episodes or 1 Jan 2015 to 30 Jan 2015) and testing data $\mathcal{B}_{31}^{60}$ (data from \{31, \dots, 60\} episodes).

\paragraph{\begin{experiment}\label{exp:crdassignment} Credit Assignment:\end{experiment}}
We define another experiment to investigate the impact of the cost definition on the training and performance of the learned optimum policy to answer~\qref{q:parametersettings} (What is the impact of varying parameter settings during training on the performance of RL agent?). Credit is assigned to each transition (taking an action on a given state) based on the defined cost function in the MDP formulation. Based on the different cost functions defined in~\sref{sec:costfun}, we generate 3 different experience sets. 
summarized in~\tref{table:CAExpSets}. Each experience set is used to train a control policy, for example, RL$_{q}$ is a trained control policy from experience set $\mathcal{F}_5$ which uses a quadratic cost function. 

In the case of different cost functions, the space complexity of the experience set ($\mathcal{F}$) is not affected. Linear cost functions are dependent on the optimal charging policies for the preceding $E \in \{1,5,10\}$ days (\eref{eq:RL_linavgcost} and \eref{eq:RL_linmedcost}). 
To train the policies on the same size of data, but with different preceding days, we evaluate the cost functions using a rolling window validation where the size of training sets is kept fixed (\fref{fig:rollingwindow}). 
Data from used for weekdays, which have similar {\EV} session characteristics (shown previously in~\cite{Manu_2020_Energies}). The training dataset has data from weekdays of 90 episodes, the testing dataset has weekdays of 30 episodes, and the rolling window comprises 30 episodes. 

\subsection{Performance Evaluation}
\label{sec:evaluvation}

\paragraph{Policy Performance:} 
To evaluate the performance of the learned policy, we use a metric defined as \textit{normalized load},
which is relative to the load achieved by the optimal policy (obtained from formulating the load flattening problem as a quadratic optimization problem). The quadratic load incurred in an episode $e$ by using policy {$\pi$} is defined by~\eref{eq:policyload}.
The normalized load for episode $e$ is calculated using~\eref{eq:RL:normalizedload}. If the trained control policy $\pi$ reaches the optimal policy, then $\tilde{L}_{\pi}^{e}$ = 1. 
\begin{equation}
    L_{\pi}^{e} = \sum\limits_{T} \left( \sum\limits^{S_{max}-1}_{d=0} N_d \cdot \textbf{u}_{s}^d   \right)^{2}
\label{eq:policyload} 
\end{equation}
\begin{equation}\label{eq:RL:normalizedload}
\tilde{L}_{\pi}^{e}=\frac{L_{\pi}^{e}}{L_{\text{opt}}^{e}}
\end{equation} 

Furthermore, for performance comparison we include the normalized load for 
\begin{romanlist}
 \item BAU: a business-as-usual (BAU) policy characterized by continuously charging each {\EV} upon arrival, and
 \item Heur: a discrete-action heuristic policy that assumes that individual EVs are charged uniformally over their entire connection time.\footnote{Specifically, the heuristic spreads the $c$ slots during which the EV needs to charge over the total number of available slots $d$. This amounts to distributing $d - c$ no-charge slots evenly over the total number of $d$ slots, thus splitting them into $d - c + 1$ parts. Assuming for simplicity that $c \geq d/2$, this means we insert a no-charge slot every $\floor{d / (d - c + 1)}$ other slots.}
\end{romanlist}

\paragraph{Policy Training Time:} 
Defined as the time it takes for RL agent to be trained. During FQI, we run 12 iterations to train the ANN for each selected training dataset. We record the time for these iterations for all the learned policies. 
 
\section[Experimental Results]{Experimental Results}
\label{sec:results}

In this section, we provide the results from~\expref{exp:obs} and~\expref{exp:crdassignment} to answer~\qref{q:impactofmdp}-\qref{q:parametersettings}. More specifically, we analyze the performance and training time of control policies trained on different $state$-$action$ representations (\expref{exp:obs}) and different cost functions (\expref{exp:crdassignment}).

\subsection{Observability: State-Action Representation}
\label{sec:results_obs}

\begin{figure*}[pos = !tb, ,align=\raggedright]
\centering
  \includegraphics[width=0.99\linewidth]{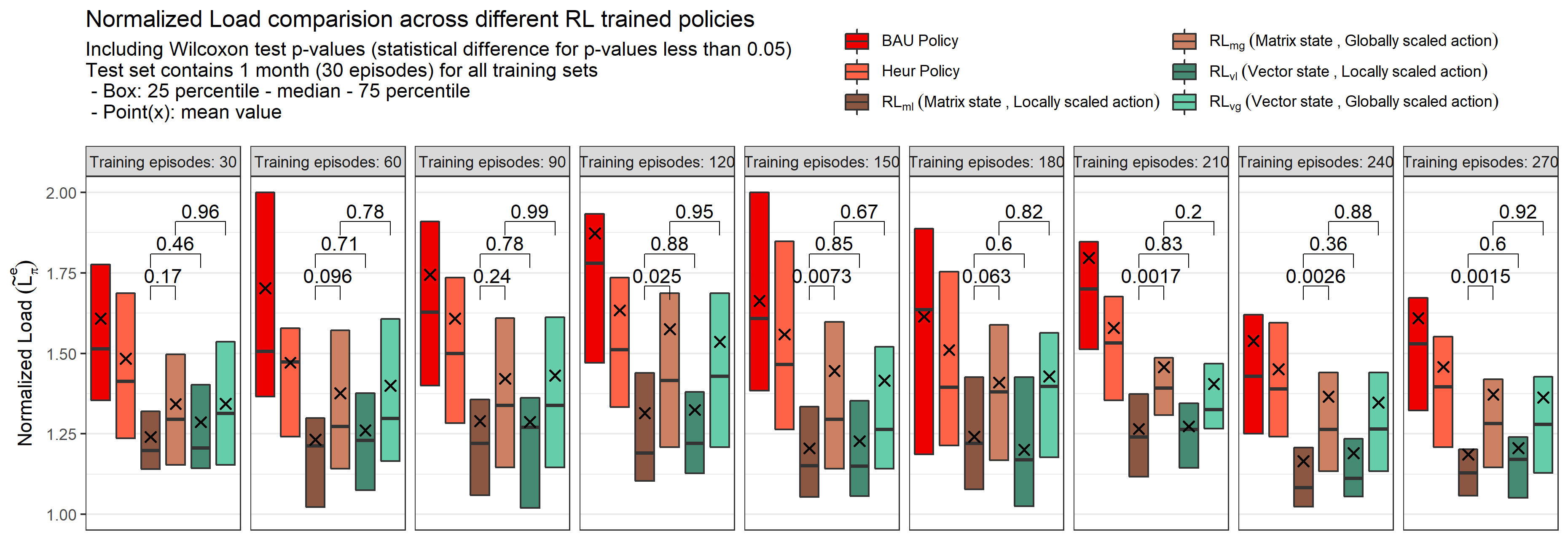}
  \caption{Normalized load for RL based control policies trained with different $state$-$action$ representations. Each box is constructed from normalized loads of 30 episodes in the test set. (Wilcoxon test p-values: statistically significant difference for p-value < 0.05) }
  \label{fig:Load_SA_compare}
\end{figure*}
\begin{figure*}[pos = !tb, ,align=\raggedright]
\centering
  \includegraphics[width=0.99\linewidth]{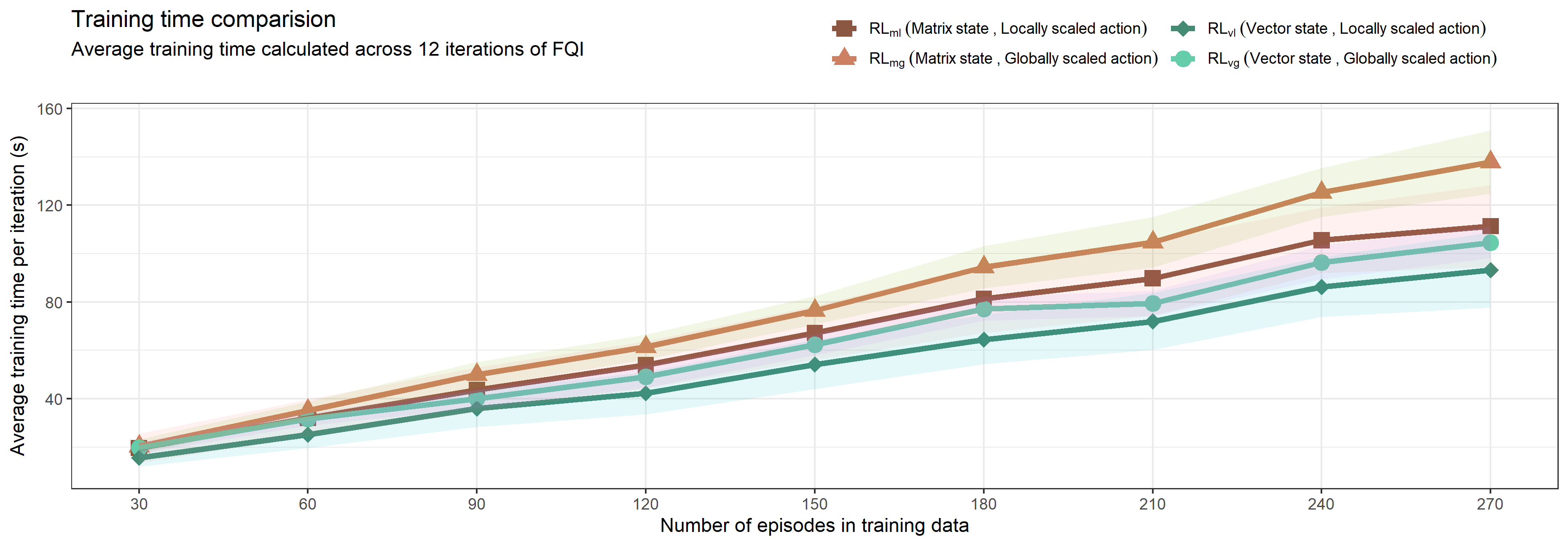}
  \caption{Training time for RL based control policies trained with different $state$-$action$ representations. (Points and solid lines: Average value calculated across all validation sets, Shaded area: 25 to 75 percentile)}
  \label{fig:Time_SA_compare}
\end{figure*}

We evaluate policies trained on different $state$-$action$ representations by analyzing the results from~\expref{exp:obs} (\sref{sec:Experiments}). \Fref{fig:Load_SA_compare} provides the normalized load ($\tilde{L}_{\pi}^{e}$) comparison for control policies for different MDPs. 
Each box is constructed from 30 normalized loads ($\tilde{L}_{\pi}^{e}$) calculated for each episode in the test set. We also perform a Wilcoxon signed-rank test on these normalized loads to quantify statistically significant difference among different control policies (significant for p-values $\leq$ 0.05). Using RL based demand coordination provides 30\%-50\% improvement in performance compared to the BAU control policy, depending on the training set size and the underlying $state$-$action$ representation in the MDP formulation.

Locally scaled actions are scaled with {\Nflex}, \ie the number {\EV}s in flexibility bin {\tflex} = $d$, in contrast to globally scaled actions which are scaled with {\Nmax}, \ie maximum system capacity (\sref{sec:action}). Locally scaled actions train a better performing control policy compared to globally scaled actions. The reason is that locally scaled actions explicitly calculate percentage of {\EV}s in each flexibility bin, which helps in learning a superior policy, in contrast to globally scaled actions where this information is implicit.
This performance gain increases with the increase in training data size. For a provided action representation, both matrix and vector state representations have similar performance. Furthermore, the performance of the control policies improves with an increase in the training data size, where we see that {$RL_{vl}$} (vector state, locally scaled actions) and {$RL_{ml}$} (matrix state, locally scaled actions) outperform the other policies.

Training time depends on the MDP formulation as each $state$-$action$ representation has different space complexity. Utilizing vector state representation in MDP formulation leads to linear space complexity \ie $\mathcal{O}(S_{max})$, compared to the quadratic space complexity in matrix states representations, \ie $\mathcal{O}(S_{max}^2)$  (\tref{table:SArepExpsets}). This results in a reduction in training time of control policies trained using vector state representations compared to matrix state representations, as shown in~\fref{fig:Time_SA_compare}.
Training time increases with the number of episodes in the training data. We note that local scaling of actions decreases the training time compared to global scaling, and the lowest training times are reported for  $RL_{vl}$ (vector state, locally scaled actions). The training time for $RL_{vl}$ policy is 30\% less than the training time for $RL_{mg}$. 

Policies learned from vector state representations perform similar to the policies learned from matrix state representations. Thus, information about flexibility (vector state representation) is enough to train a good policy, which answers~\qref{q:stateactionfeatures} (What features should be used to define $state$-$action$ representations?).
Furthermore, training a control policy on a partially observable MDP (such as a MDP based on vector state representation and locally scaled actions) takes approximately 20\%-30\% less time compared to the fully observable MDP. This answers~\qref{q:impactofmdp} (effect of different MDP formulations on RL policy?), where we note that policies trained using MDPs based on vector states and locally scaled actions outperform other polices, are space-efficient, and are computationally less expensive. Furthermore, we notice that performance of RL policies increases with increase in the time span of training data, which answers a part of~\qref{q:parametersettings} (impact of varying parameter settings?).

\subsection{Impact of Cost Functions Definitions}
\label{sec:results_crd}

\begin{figure*}[pos = !tb, ,align=\raggedright]
\centering
\begin{subfigure}[t]{.48\textwidth}
  \centering
  \includegraphics[width=0.99\linewidth]{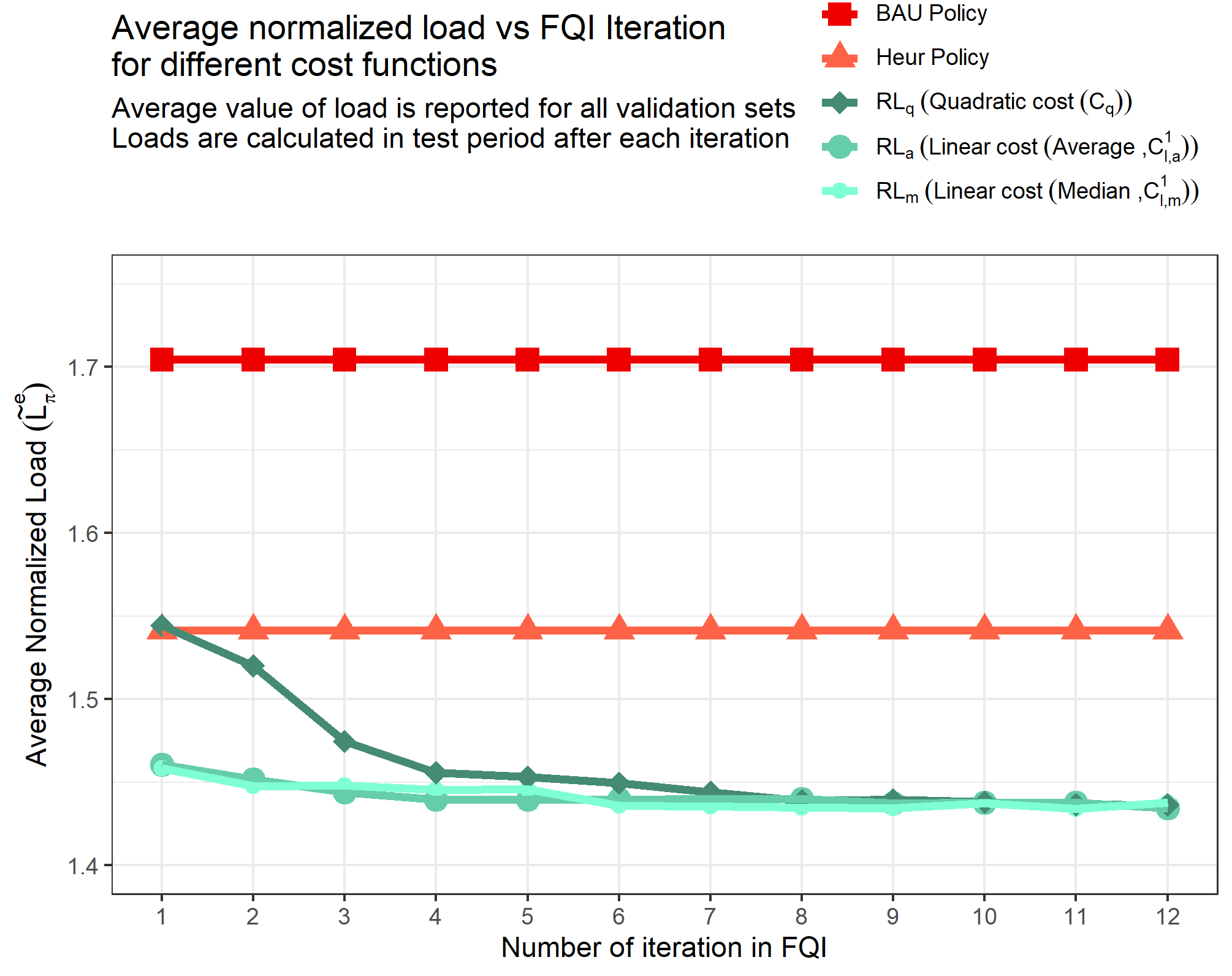}
  \caption{Average normalized load per iteration. Linear cost functions with $E$ = 1, \ie optimum solution of one preceding day ($C_{l,a}^1$ and $C_{l,m}^1$ ). (Average of all episodes and all validation test sets)}
  \label{fig:load_costfuns}
\end{subfigure}%
\hfill
\begin{subfigure}[t]{.48\textwidth}
  \centering
  \includegraphics[width=0.99\linewidth]{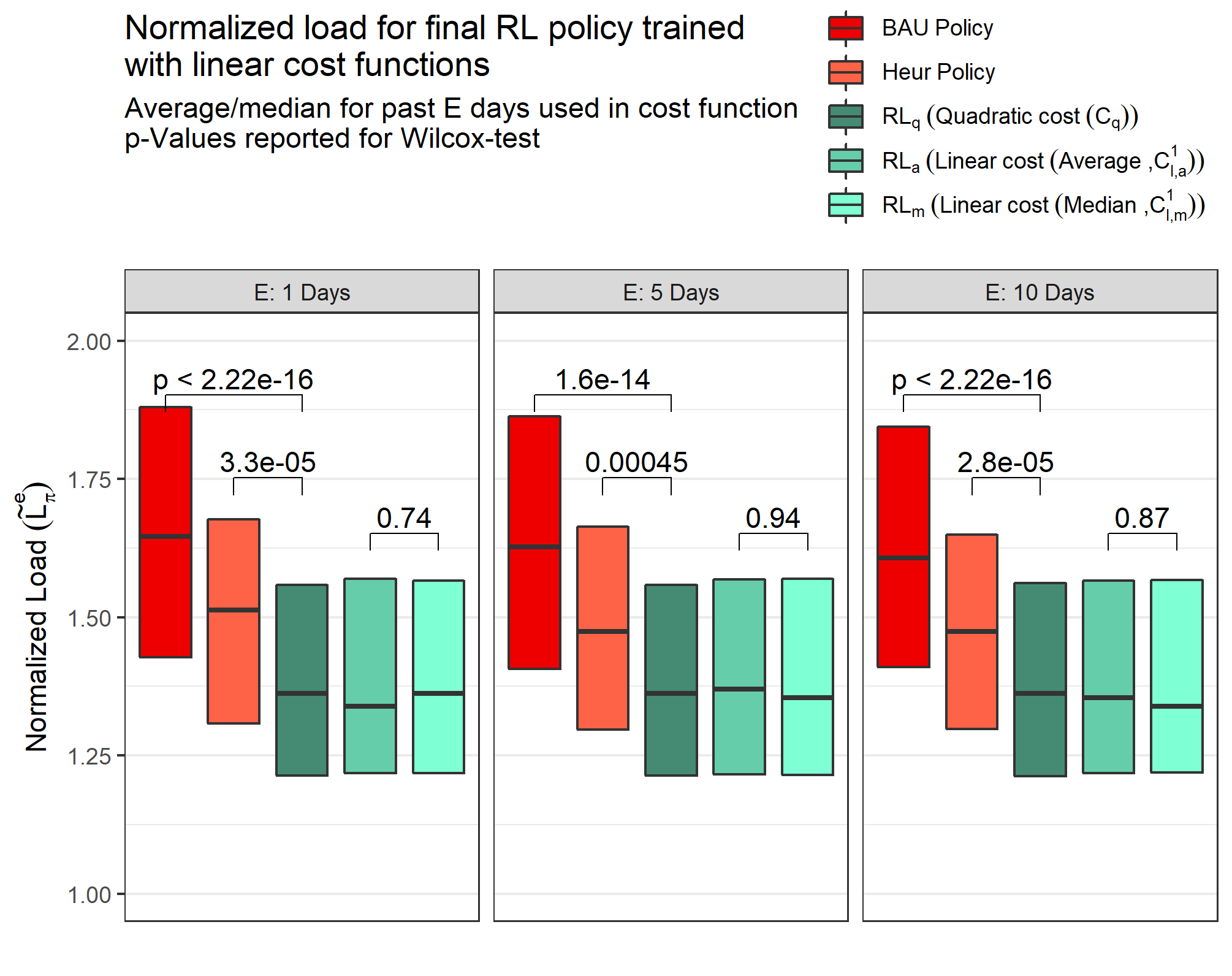}
  \caption{Normalized load for RL based control policies with linear cost functions that use $E \in \{1,5,10\}$. (Wilcoxon test p-values: statistically significant different for p-value<0.05) }
  \label{fig:load_linearcost}
\end{subfigure}
\caption{Results for RL based control policies trained with different cost functions}
\label{fig:creditassignment}
\end{figure*}

The performance of a trained control policy depends on the cost function, where an effective cost function helps to achieve faster convergence by providing informative rewards. To study the effect of cost functions on performance and convergence, we analyze the results of~\expref{exp:crdassignment} (\sref{sec:Experiments}). We use the FQI algorithm to train control policies using MDP formulations characterized by the different cost functions defined in~\sref{sec:costfun}.
\fref{fig:load_costfuns} compares the average normalized load incurred in different control policies. Control policies are evaluated in the test set after each iteration, and the average normalized load is calculated for all episodes in all validation sets. Note that linear cost functions ({\CRLAvg} and {\CRLMed}) are based on the optimum solutions (power consumption with optimum policy {\EV} charging demand coordination) of preceding $E$ episodes, and we choose $E = 1$ for this comparison. Similar results are observed for other values of $E$. Furthermore, we also include the average normalized load incurred by a BAU control policy for comparison.

Policy based on quadratic cost function ($RL_{q}$) takes 8-10 iterations of FQI algorithm to converge, whereas, the policies trained on linear cost functions ($RL_{a}$ and $RL_{m}$) take 3-4 iterations to converge. FQI takes longer to converge with the quadratic cost function because the agent has to learn the expected T-step return. This information is indirectly included in the linear cost functions, where we include the optimum policies of preceding days. Furthermore, we notice after a single iteration, both control policies based on linear cost functions perform much better than the policy trained with a quadratic cost function. The RL agent in our case has to learned the 12-step return starting from the root node (based on {\Smax} = 12). For an $N$-step return ($N\gg12$), a policy trained with the quadratic cost function will take a large number of iterations to converge, whereas we can train a similar policy in fewer iterations using a linear cost function. 

\fref{fig:load_linearcost} shows the normalized load for control policies trained for linear cost functions with $E \in \{1, 5, 10\}$ (average/median of optimum solutions is calculated for preceding $E$ episodes). We also include the p-values for the Wilcoxon signed-rank test to evaluate statistical difference (statistically significant difference for p-values $\leq$ 0.05). We notice similar performance for average ({\CRLAvg}) and median ({\CRLMed}) based linear cost functions. Policy trained with average/median of preceding 10 episodes has similar performance to a policy trained with information of a single preceding episode.

FQI converges faster with linear cost functions because they are easier to learn and are based on the optimum policies of preceding episodes. This helps in answering~\qref{q:costfunctiondef} (Impact of the cost function definition on the learning speed and performance of the learned policy?), as we can conclude from~\fref{fig:load_costfuns} that we should utilize the optimum policies of preceding days to define linear cost functions, which prove to be superior to the quadratic cost function. We also answer~\qref{q:parametersettings} (impact of varying parameter settings?) in this section, where we note from~\fref{fig:load_linearcost} that the optimum solution of the immediate preceding episode ($E$ = 1) is as informative as the average of preceding 10 episodes.

\section[Conclusions]{Conclusions}
\label{sec:conclusions}



In this paper, we explored different aspects of an MDP formulation used to train a reinforcement learning based charging demand coordination to jointly coordinate  the charging demand of a group of {\EV} charging stations. We defined 
\begin{romanlist}
    \item a new partially observable state representation,
    \item different action representations, and
    \item two new cost functions that assign costs based on a posteriori determined optimal charging actions for the preceding days 
\end{romanlist}. 
These $state$-$action$ representations and cost functions are used to define different MDP formulations. 
A real-world {\EV} charging dataset is used to evaluate the performance of RL based control policies trained using different MDP formulations. We compared our control policies with an optimum policy (an all-knowing policy), a business-as-usual (BAU) control policy (\ie charging fully upon arrival) and a heuristic control policy (spreading out each individual EV's required charging over the time it is connected). The conclusions from our study can be summarized as follows:
\begin{enumerate}[label={(\arabic*)}]
  \item \textit{\textbf{State-Action Representation}}: 
  A partially observable vector $state$ representation provides similar performance to a fully observable matrix $state$ across all training set sizes (\sref{sec:results_obs}). Information about flexibility is sufficient to learn a control policy, as vector state representations are based on aggregate flexibility. Furthermore, we see that local scaling of actions (scaling by the number of {\EV}s with the same flexibility, {$N_{d}$}) improves the performance of control policy compared to full scaling (scaling by the system capacity, {\Nmax}).
  We conclude that our vector state representation and locally scaled actions are the best choice for $state$-$action$ representation among the considered alternatives.
  
  \item \textit{\textbf{Cost Function}}: 
  The proposed linear cost functions expedite optimization compared to quadratic cost function by,
  \begin{romanlist}
      \item learning a control policy with 10\% lower normalized load in a single iteration (\sref{sec:results_crd}), and
      \item achieving convergence within fewer iterations as compared to the quadratic cost function
  \end{romanlist} (\fref{fig:load_costfuns}).
  Furthermore, the number of preceding days utilized in linear cost functions does not affect the performance significantly, thus, the optimum solution from a single preceding day is sufficient (\fref{fig:creditassignment}). 
  
  \item \textit{\textbf{Space and Computational Complexity}}: 
  MDP formulations based on vector states and matrix states have linear and quadratic space complexity respectively, in terms of the total number of decision slots, \ie {\Smax} (\sref{sec:Experiments}). Vector state representation and locally scaled actions assist in learning control policies 30\% faster compared to all other alternatives (\fref{fig:Time_SA_compare}).  
\end{enumerate}

An MDP formulation based on
\begin{romanlist}
    \item a partially observable vector $state$ representation,
    \item a locally scaled $action$ representation, and
    \item a linear cost function
\end{romanlist}, 
can provide 40\%-50\% improvement in load flattening compared to the BAU control policy and 20\%-30\% compared to a simple heuristic control policy for {\EV} coordination (\fref{fig:Load_SA_compare} and \fref{fig:load_linearcost}). 

In future work, we will
\begin{romanlist}
    \item explore possible improvements in the function approximator used to approximate the $Q$-function in the FQI algorithm. The currently used Artificial Neural Network (ANN) can be improved by hyperparameter tuning or using a different neural network architecture. Further, 
    \item we currently use a value iteration based learning approach to train the RL agent, where we learn the $Q$-function based on discrete state-action pairs. We would like to investigate policy iteration methods that learn the policy directly.
\end{romanlist}

\section*{Declaration of Competing Interest}

{The authors declare that they have no known competing financial interests or personal relationships that could have appeared to influence the work reported in this paper.}

\section*{Acknowledgements}

{
Part of this research has received funding from the European Union's Horizon~2020 research and innovation programme for the projects BRIGHT (\url{https://www.brightproject.eu/} -- grant agreement no. 957816) and RENergetic (\url{https://www.renergetic.eu/} -- grant agreement no. 957845).}

\bibliographystyle{elsarticle-num-names}
\bibliography{refs/references, refs/references-own}

\printcredits


\end{document}